\documentclass[runningheads]{llncs}
\usepackage[T1]{fontenc}
\usepackage{graphicx}
\usepackage{amsmath}
\usepackage{amssymb}
\usepackage{subcaption}
\usepackage{booktabs}
\usepackage{gensymb}
\usepackage{enumitem}
\usepackage[normalem]{ulem}

\usepackage{xcolor}

\newcommand{\be}[1]{#1}

\newcommand{\matsup}[2]{\mathbf{#1}^\mathrm{#2}}

\newcommand{\defeq}{\mathrel{\mathop:}=}

\begin{document}
\title{Latent attention on masked patches \\ for flow reconstruction}
%
\author{Ben Eze\inst{1}\orcidID{0009-0009-8594-082X} \and
Luca Magri\inst{1,2}\orcidID{0000-0002-0657-2611} \and
Andrea Nóvoa\inst{1,3}\orcidID{0000-0003-0597-8326}}
\authorrunning{Eze et al.}
%
\institute{
Aeronautics Dept., Imperial College London, Exhibition Rd, London SW7 2BX, UK\and
DIMEAS, Politecnico di Torino, Corso Duca degli Abruzzi, 24, Torino, 10129, Italy
\and
I-X, Imperial College London, 84 Wood Lane, London W12 0BZ, UK \\
\email{\{ben.eze21,l.magri,a.novoa\}@imperial.ac.uk}}
\maketitle              
\begin{abstract}
Vision transformers have shown outstanding performance in image generation, yet their adoption in fluid dynamics remains limited. 
We introduce the \textit{Latent Attention on Masked Patches} (LAMP) model, an interpretable regression-based modified vision transformer designed for masked flow reconstruction. 
LAMP follows a three-fold strategy: 
(i) partition of each flow snapshot into patches, 
(ii) patch-wise dimensionality reduction via proper orthogonal decomposition, and 
(iii) reconstruction of the full field from a masked input using a single-layer transformer trained via closed-form linear regression. 
We test the method on two canonical 2D unsteady wakes: a laminar wake past a bluff body, and a chaotic wake past two cylinders.
On the laminar case, LAMP accurately reconstructs the full flow field from a 90\%-masked and noisy input, across signal-to-noise ratios between 10 and 30\,dB. 
Further, the learned attention matrix yields interpretable multi-fidelity optimal sensor-placement maps.  
LAMP's performance on the chaotic wake is limited, but outperforms other regression methods such as gappy POD.  
The modularity of the framework, however, naturally accommodates nonlinear compression and deep attention blocks, thereby providing an efficient baseline for nonlinear, high-dimensional masked flow reconstruction.

\end{abstract}

\section{Introduction}

Reconstructing high-dimensional fields from sparse, noisy, and masked measurements is a central challenge in scientific machine learning and fluid mechanics~\cite{Lam2023,Xia2024,Mo_Magri_2026}. 
In experimental settings, particle image velocimetry (PIV) measurements cover only a subset of the domain, and, since each run captures an independent realisation of the flow, PIV is often limited to mean-flow analyses rather than time-resolved full-field measurements in high-dimensional settings, e.g.,~\cite{bekoglu2025formation}. 
To reconstruct full flow fields, data-driven approaches have been employed, such as convolutional neural networks~\cite{Lee2017,Yaxin2025}, generative adversarial networks~\cite{nista2024influence}, or graph neural networks~\cite{Quattromini2025}, among many others. 
However, these methods typically assume that a large fraction of the field is observed and are primarily designed for super-resolution, in-painting, or mean-flow reconstruction tasks.
Real-time full-field reconstruction from sparse measurements can alternatively be achieved by coupling reduced-order models with data assimilation~\cite{cheng2024efficient,Ozalp2025}.
Vision Transformers (ViTs)~\cite{Dosovitskiy2021}, in particular, have become the state-of-the-art backbone for image generation and large language models~\cite{Peebles2023}, but have not been fully  adopted yet in the fluid dynamics community.
Here, we introduce the \textit{Latent Attention on Masked Patches} (LAMP) model, inspired by ViTs and masked autoencoders~\cite{He2022}. 
The key idea is to 
(i) partition each flow snapshot into non-overlapping patches; 
(ii) compress each patch into a low-dimensional latent representation; here,  patch-wise Proper Orthogonal Decomposition (POD)~\cite{Berkooz1993}; and 
(iii) reconstruct the full field from a subset of the patches, i.e., a masked input, using an attention-based patch-to-patch prediction mechanism; here, implemented as a single-layer transformer trained via closed-form linear regression. 
This design guarantees convergence to a global minimum, eliminates dependence on learning-rate schedules and random weight initialisation, and yields fast, reproducible and physically interpretable results. 
Further, the modular architecture allows expressive components to be incorporated where performance demands it---such as nonlinear autoencoders~\cite{Fukami2019,Doan2023} or multi-layer transformers.
The paper is structured as follows. 
Section~\ref{sec:method} details the proposed LAMP. 
Section~\ref{sec:results} demonstrates LAMP on two canonical unsteady wakes: a laminar bluff-body wake and a chaotic wake over two adjacent cylinders. 
Section~\ref{sec:conclusions} closes the paper with conclusions and future work directions.
\section{Latent Attention on Masked Patches (LAMP)}~\label{sec:method}
\begin{figure}[!ht]
    \centering
    \includegraphics[width=0.9\linewidth]{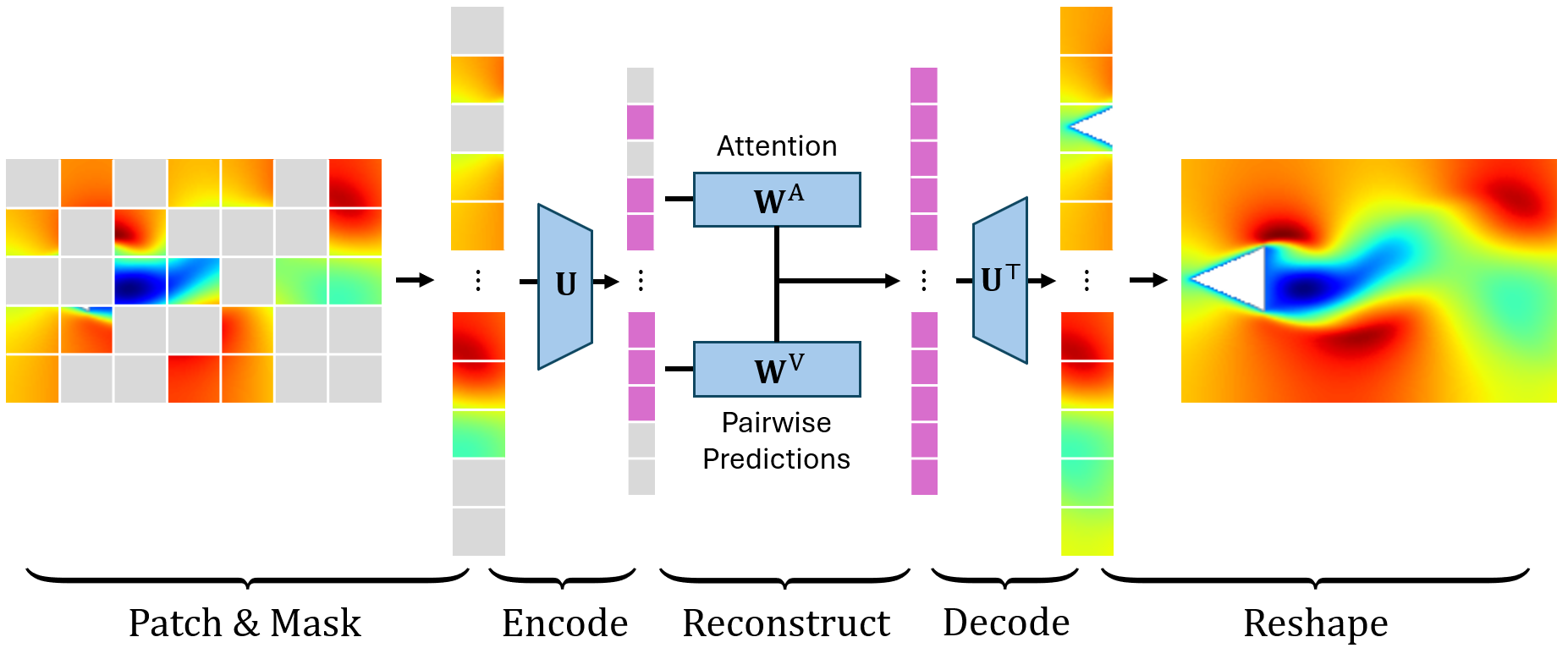}
    \caption{Pictorial illustration of the patch-wise latent attention model (LAMP). A linear autoencoder embeds the input in a lower-dimensional latent space, where predictions can be made efficiently.  The predictions are pair-wise (i.e.
    each patch predicts every other patch) and weighted with a confidence score.}
    \label{fig:rViT_architecture}
\end{figure}

As illustrated in Fig.~\ref{fig:rViT_architecture}, LAMP  consists of three stages: 
(i) data patching and reshaping; 
(ii) patch-wise dimensionality reduction via linear autoencoding with proper orthogonal decomposition (POD);  and
(iii) attention-based patch-to-patch prediction in the latent space.

 \begin{enumerate}[label={}, leftmargin=0pt, itemsep=4pt]
\item \textbf{Patching. } 
{ First, we divide the normalized input 2D data (of height $H$, width $W$ and $C$ components) $\mathbf{X}_\textrm{in}\in\mathbb{R}^{H\times W \times C}$ into $N$ size-$P$ square patches of dimension $D \defeq CP^2$, which are flattened to obtain the patched data  $\mathbf{X}\in \mathbb{R}^{N \times D}$ (see \cite{Dosovitskiy2021}). 
}
\item 
\textbf{Patch-wise POD.} 
{
We apply dimensionality reduction to compress  $\mathbf{X}$ into $\mathbf{Z}\in\mathbb{R}^{N\times N_e}$, with embedding (latent) dimension $N_e \ll D$.    
Each patch vector $\mathbf{x}_n\in\mathbb{R}^{D}$ is compressed into $\mathbf{z}_n\in\mathbb{R}^{N_e}$ by minimizing 
\begin{equation} \label{eqn:loss_AE}
            \matsup{\mathcal{L}}{AE} = \sum_{n=1}^N
            \left\lVert {\tilde{\mathbf{x}}_n- \mathbf{x}_n}\right\rVert^2_2, \quad \text{subject\,to } \;\; \tilde{\mathbf{x}}_n= \mathbf{U}_n\mathbf{U}_n^\top \mathbf{x}_n. 
        \end{equation}
The patch-wise optimal orthonormal bases $\mathbf{U}_n\in\mathbb{R}^{D\times N_e}$ are obtained from the training time series as 
$\matsup{X}{train}_n \approx \mathbf{U}_n\mathbf{\Sigma} (\matsup{V}{train}_n)^\top$, where the approximation arises as we retain only the first $N_e$ leading singular vectors.  
 Finally, we construct the encoding and decoding operators, $\mathbf{E}$ and $\mathbf{D}$, as block-diagonal matrices composed of the patch-wise bases $\{\mathbf{U}_n\}_{n=1}^N$ such that
\begin{equation} \label{eqn:linear_ROM}
\mathbf{Z} = \mathbf{E}\mathbf{X}, \quad \text{and}\quad  \tilde{\mathbf{X}} = \mathbf{D}\mathbf{Z},
\end{equation}
where the $n$-th block in $\mathbf{E}$ and $\mathbf{D}$ contain $\mathbf{U}_n^\top$ and $\mathbf{U}_n$, respectively.  %
The patched latent data $\mathbf{Z}$ is then passed to the transformer. A low reconstruction loss $\mathcal{L}^\mathrm{AE}$ is a prerequisite for accurate masked predictions.  
\item{
\textbf{Masked prediction.} 
We modify the attention equation presented in \cite{vaswani2017} for the task of single-layer reconstruction. The aim is to predict the full latent snapshot  $\mathbf{Z}^\star$ from a masked input $\matsup{Z}{masked}$, where ${Z}^\mathrm{masked}_n=0$ for all masked indices $n \in \mathcal{N}_{masked}$.
Each component in $\mathbf{Z}^\star$ is 
\begin{align} \label{eqn:masked_reconstruction}
    {Z}^\star_{me} = 
    \sum_{n=1}^{N} \mathrm{softmax}(\mathbf{a}_m)_n 
    {V}_{mne}, \quad \text{for } \left\{\begin{array}{cc}
         m = 1,\dots,N,\\ e = 1,\dots,N_e 
    \end{array}\right.
\end{align}
where $\mathbf{a}_m$ is the $m$-th row of the log-attention matrix $\mathbf{A}\in\mathbb{R}^{N\times N}$, and  $\mathbf{V}\in\mathbb{R}^{N\times N\times N_e}$ is the value tensor; these are defined as 
\begin{align}
    {A}_{mn} = \sum_{e=1}^{N_e} {W}^\mathrm{A}_{mne}\, Z^{\text{masked}}_{ne}, \quad\text{and}\;\; {V}_{mne} = \sum_{f=1}^{N_e} {W}^\mathrm{V}_{mnef}\, Z^{\text{masked}}_{nf}, 
\end{align}
where 
$\matsup{W}{V} \in \mathbb{R}^{N \times N \times N_e \times N_e}$ and $\matsup{W}{A}\in \mathbb{R}^{N \times N \times N_e}$ are the value and attention weight-tensors, which are trained on \textit{unmasked} data with a two-step approach: 
\begin{enumerate}[label=\arabic*.]
\item The value tensor $\matsup{W}{V}$ can be viewed as an $N \times N$ grid of matrices:
    each block $\matsup{W}{V}_{mn} \in \mathbb{R}^{N_e \times N_e}$ defines a
    linear map that predicts patch $\mathbf{z}_m$ from patch $\mathbf{z}_n$. 
    These blocks are obtained by least-squares regression,
\begin{subequations} 
\begin{equation} \label{eqn:finding_WV}
    (\matsup{Z}{train})_m \approx \matsup{W}{V}_{mn} (\matsup{Z}{train})_n,
\end{equation}
where the corresponding linear prediction error is 
        \begin{equation} \label{eqn:loss_lin}
            \matsup{\mathcal{L}}{lin}_{mn}=
            \left\lVert(\matsup{Z}{train})_m - 
            \matsup{W}{V}_{mn} (\matsup{Z}{train})_n\right\rVert^2_2. 
        \end{equation}
\end{subequations}
\item The attention tensor $\matsup{W}{A}$ is learned in a second stage and can
    be decomposed analogously, as a $N \times N$ grid of vectors $w^\mathrm{A}_{mn} \in \mathbb{R}^{N_e}$. The error $\matsup{\mathcal{L}}{lin}_{mn}$
    provides a prediction uncertainty measure, so that predictions with lower
    $\matsup{\mathcal{L}}{lin}_{mn}$ are assigned higher attention weights. In practice, we
    regress the \textit{log-error},
    \begin{equation}\label{eqn:finding_WA}
    -\log(\matsup{\mathcal{L}}{lin}_{mn})\approx \matsup{W}{A}_{mn} (\matsup{Z}{train})_n. 
    \end{equation}
\end{enumerate}
}
At inference, the log-error is set to $-\infty$ for masked patches, so that $\mathrm{softmax}$ assigns them zero attention weight.  
The model is then evaluated on unseen test data $\matsup{X}{test}$. The overall prediction error is defined as
\begin{equation} \label{eqn:loss_pred}
    \matsup{\mathcal{L}}{pred}=
    \big\| {\tilde{\mathbf{X}}^\star - \mathbf{X}^\mathrm{test}}\big\|^2_2  =
    \big\| {\mathbf{D}{\mathbf{Z}}^\star - \matsup{X}{test}}\big\|^2_2. 
\end{equation}
}
\end{enumerate}
\section{Results}\label{sec:results}

We deploy LAMP to reconstruct the streamwise $u$ and spanwise $v$ velocity fields of two 2D unsteady wakes. 
First, we test LAMP on the laminar wake of a triangular bluff body at Reynolds number $Re=100$~\cite{Yaxin2025} in \S\ref{sec:results-laminar} and analyse the effect of varying latent-space dimension, noise intensity, and patch placement. 
We use 100 snapshots  for training and 60 for testing (one shedding period is $32$ snapshots).  
Second, \S\ref{sec:results-chaotic} showcases LAMP on a chaotic wake past two cylinders of diameter $d$ with centres spaced $1.5d$ apart vertically. The flow has $Re=200$, Mach number $M=0.1$,  uniform inlet velocity, convective outflow, and periodic side boundary conditions. 
After discarding transients, 7000 snapshots ($\Delta t = 1.39$) are  split 75\%/20\% for training and testing, discarding the intermediate 5\% to reduce temporal leakage. Both datasets are standardized to zero mean and unit variance for a fair comparison between them.

\subsection{Laminar wake results}\label{sec:results-laminar}

\begin{figure}[!h]
    \centering
    \includegraphics[width=.99\textwidth]{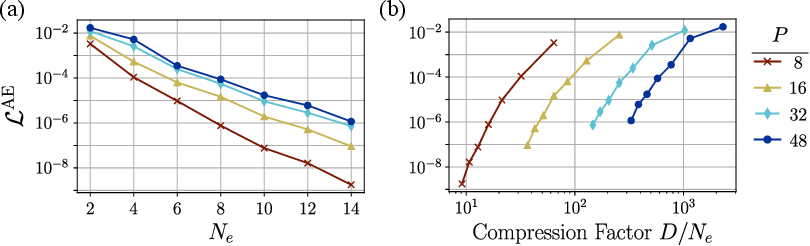}  
    \caption{
    Patch-wise POD reconstruction performance for varying patch size $P$.  Loss $\mathcal{L}^\mathrm{AE}$ versus (a) latent dimension $N_e$ and (b) compression factor $D/N_e$. 
    }
    \label{fig:AE_recon_MSE}
\end{figure}

First, we validate the patch-wise POD and examine the loss  $\mathcal{L}^\mathrm{AE}$ (\ref{eqn:loss_AE}), which sets the lower bound for the masked reconstruction error. 
Figure~\ref{fig:AE_recon_MSE} shows that 
(i) $\mathcal{L}^\mathrm{AE}$ saturates rapidly with $N_e$, which is consistent with the well-known energy concentration of laminar bluff-body wakes into a few dominant POD modes; and 
(ii) larger patches reach the same $\mathcal{L}^\mathrm{AE}$ at higher compression factors, which indicates that using very small patches may fragment the coherent vortex structures into features that  carry little dynamical information. 

\begin{figure}[!hbt]
    \centering
    \includegraphics[width=.99\linewidth]{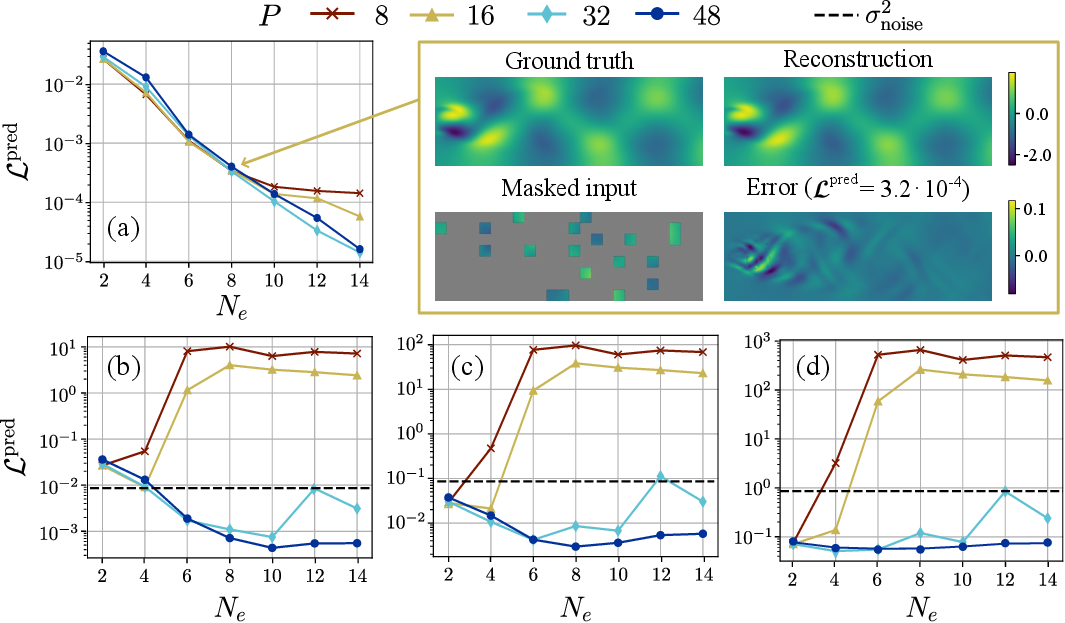}
    \caption{Reconstruction error $\matsup{\mathcal{L}}{pred}$ for varying noise levels, patch size $P$ and latent dimension $N_e$, \be{all with 10\% patches unmasked}. 
    The input is noisy masked data with SNR = (a) $\infty$ (noise-free), (b) 30 dB, (c) 20 dB, (d) 10 dB.  The loss is computed against the noise-free target $\matsup{X}{test}$ \be{and the model is trained on noise-free data}.  Horizontal dashed lines show the noise variance. $\mathcal{L}^\mathrm{pred}$ is calculated for 25 random patch arrangements and the \textit{median} is plotted.
    An example reconstruction  for the noise-free case with $P=16$, $N_e=8$ is shown. 
    }
    \label{fig:masked_prediction_MSE}
\end{figure}

Figure~\ref{fig:masked_prediction_MSE} shows the LAMP masked flow reconstruction results from randomly unmasked patches covering 10\% of the domain for noise-free  (Fig.~\ref{fig:masked_prediction_MSE}a) and noisy input  (Fig.~\ref{fig:masked_prediction_MSE}b-c). 
Noise is added to the input as 
$\matsup{\tilde{X}}{noisy}=\matsup{\tilde{X}}{masked}+\boldsymbol{\varepsilon}$, 
where $\tilde{(\cdot)}$ denotes unnormalized data, 
$ \boldsymbol{\varepsilon} \sim \mathcal{N}(0,\mathbf{\Sigma})$, and 
$\mathbf{\Sigma}=\mathbb{E}[\matsup{\tilde{X}}{masked}]{10^{-\mathrm{SNR}/10}}$.

In the noisy cases, the  LAMP provides flow-field reconstructions with prediction errors $\mathcal{L}^\mathrm{pred}$ consistently lower than the variance of the added noise. 
Increasing the latent dimension $N_e$ reduces $\mathcal{L}^\mathrm{pred}$ (Fig.~\ref{fig:masked_prediction_MSE}a), with larger patches outperforming smaller ones once sufficient modes are retained to resolve the coherent shedding.
\be{However, adding Gaussian noise shifts the $\mathcal{L}^{\mathrm{pred}}$ minima toward lower $N_e$ (Fig.~\ref{fig:masked_prediction_MSE}b-d).} 
Small patches are less robust because their local POD bases amplify sub-patch-scale noise.  In contrast, larger patches act as spatial filters that preserve the dominant flow structures. Notably, for moderate SNR a broad range of $(P,N_e)$ yields $\mathcal{L}^{\mathrm{pred}}$ below the noise variance $\sigma^2_{\mathrm{noise}}$ (horizontal dashed lines). 
Therefore, the proposed LAMP successfully achieves simultaneous noise removal and flow reconstruction from masked data.  

\begin{figure}[!h]
    \centering
    \includegraphics[width=.95\linewidth]{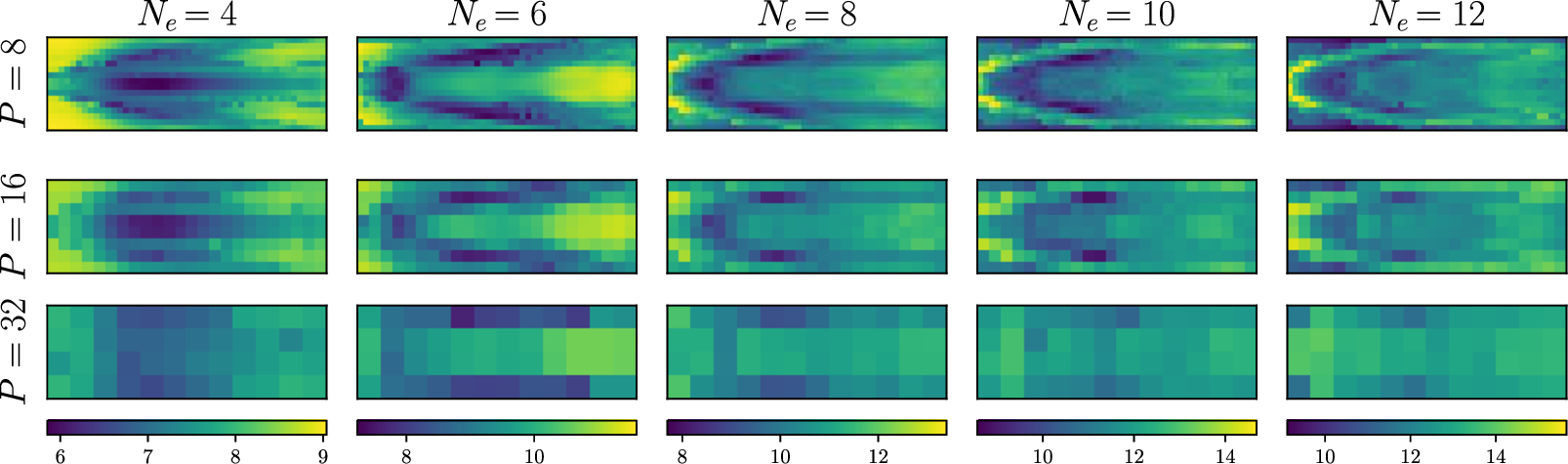}
    \caption{Predictive-power maps for varying $P$ and $N_{e}$. Patches with higher predictive power predict the rest of the patches with lower average loss.
    }
    \label{fig:predictive_power}
\end{figure}

Next, we focus on the attention matrix, which yields as a by-product an interpretable method for informing sensor placement: the predictive-power maps shown in Fig.~\ref{fig:predictive_power}, which are computed by averaging the columns of $-\log\mathcal{L}^{\mathrm{lin}}_{mn}$~\eqref{eqn:finding_WA} and reshaping into the physical space. 
Patches with highest predictive power are those whose local dynamics most accurately predict the rest of the field, i.e., the optimal sensor placement for reconstruction.  
Beyond reducing the resolution, increasing the patch size predominantly reduces the range of values in the attention map. By contrast, as $N_e$ increases, high-power regions multiply from 1-2 near-body patches to distributed coverage across shear layers and far wake, mirroring the POD hierarchy.
Placing sensors at persistent predictive-power peaks across $N_e$ thus provides optimal multi-fidelity sensor layouts for both coarse and fine robust reconstructions. 
\vspace{-5pt}

\subsection{Chaotic wake results}\label{sec:results-chaotic}

\begin{figure}[!h]
  \centering
    \includegraphics[width=.99\textwidth]{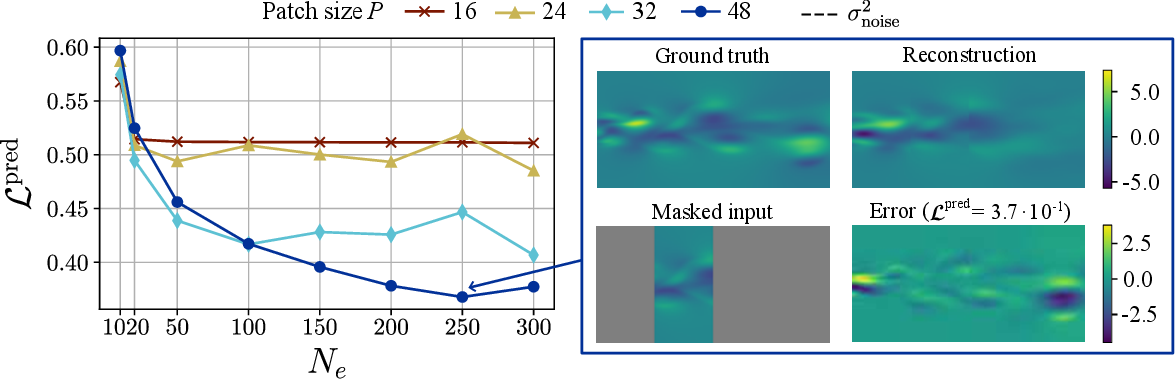}
    \caption{\be{Masked reconstruction of the chaotic wake from 25\% unmasked patches, patches placed at the locations of highest predictive power. The reconstruction of one snapshot with the best model ($P=48, N_{e}=250$) is shown on the right.}
    }     
    \label{fig:masked_recon_d2}
\end{figure}

Because of the chaotic nature of the flow,  the two-cylinder wake requires larger patches and more POD modes compared to  the laminar case, as reflected in the prediction loss in Fig.~\ref{fig:masked_recon_d2}a. 
\be{
The requirement for larger patches to minimise $\mathcal{L}^\mathrm{pred}$ indicates that each patch must span a sufficient spatial extent to capture the dominant coherent structures because the relevant dynamical correlations operate over larger spatial scales than in the laminar case. 

Despite the increased reconstruction loss compared to dataset 1, 
LAMP achieves a 26\% lower reconstruction error $\mathcal{L}^\mathrm{pred}$ than gappy POD \cite{everson1995karhunen} on identical inputs (results not shown for brevity).  
Qualitatively, gappy POD tends to overestimate velocity fluctuations in regions of greater uncertainty further downstream, whereas LAMP produces smoother reconstructions.  
LAMP's prediction accuracy is highest immediately downstream of unmasked patches and degrades further away. This is due to the spatio-temporally chaotic nature of the flow; two patches become less correlated with each other the further apart they are. 
Extending to LAMP with \textit{multi-layer} transformers is required to predict nonlinear relationships between multiple patches. Additionally, time-delay embedding \cite{sauer1991embedology} could improve accuracy by exploiting temporal correlations.
}

\section{Conclusion and future work}\label{sec:conclusions}

We introduce the \textit{Latent Attention on Masked Patches} (LAMP) model, which consists of a regression-based architecture inspired by Vision Transformers. LAMP is designed to reconstruct high-dimensional flow fields from noisy and masked measurements. 
By integrating patch-wise POD with a linearly decomposable attention mechanism, we provide a framework that avoids extensive hyperparameter tuning and back-propagation.  
We showcase the LAMP on two test cases: a laminar wake past a triangular bluff body and a chaotic wake past {two cylinders}. 
On the  laminar wake, LAMP successfully reconstructs flow fields from only 10\% unmasked patches even when the input patches are corrupted by noise levels up to $\text{SNR}=10$~dB. The reconstruction error remains an order of magnitude lower than the input noise variance, which shows that LAMP is both a reconstruction and de-noising tool. 
Beyond reconstruction, the attention maps from the LAMP provide an  interpretable multi-fidelity sensor placement approach, which  identifies the patches of highest importance for reconstruction. 
On the two-cylinder wake, LAMP outperforms gappy POD with 26\% lower $\mathcal{L}^\mathrm{pred}$. The overall error is, however, larger than the laminar case, suggesting that multi-layer attention blocks and nonlinear autoencoding might be required to reconstruct higher-dimensional, spatio-temporally chaotic flows. The modular architecture of the framework provides a natural pathway to more expressive extensions.

\begin{credits}
\subsubsection{\ackname} { LM acknowledges funding from ERC Starting Grant PhyCo 949388. AN is supported by the Eric and Wendy Schmidt AI in Science Fellowship.}
\end{credits}
\bibliographystyle{splncs04}
\bibliography{references}
\end{document}